# Peacock Exploration: A Lightweight Exploration for UAV using Control-Efficient Trajectory

EungChang Mason Lee[1][0000-0003-0219-784X], Duckyu Choi[2][0000-0002-0050-8627],

and Hyun Myung[1,2][0000-0002-5799-2026]

[1] School of Electrical Engineering, KI-AI, KI-R,
Korea Advanced Institute of Science and Technology, Daejeon, 34141, Korea
[2] Department of Civil and Environmental Engineering,
Korea Advanced Institute of Science and Technology, Daejeon, 34141, Korea
`{eungchang_mason, duckyu, hmyung}@kaist.ac.kr`
`http://urobot.kaist.ac.kr`

**Abstract.** Unmanned Aerial Vehicles have received much attention in recent years due to its wide range of applications, such as exploration of an unknown environment to acquire a 3D map without prior knowledge of it. Existing exploration methods have been largely challenged by computationally heavy probabilistic path planning. Similarly, kinodynamic constraints or proper sensors considering the payload for UAVs were not considered. In this paper, to solve those issues and to consider the limited payload and computational resource of UAVs, we propose "Peacock Exploration": A lightweight exploration method for UAVs using precomputed minimum snap trajectories which look like a peacock's tail. Using the widely known, control efficient minimum snap trajectories and OctoMap, the UAV equipped with a RGB-D camera can explore unknown 3D environments without any prior knowledge or human-guidance with only $O(\log N)$ computational complexity. It also adopts the receding horizon approach and simple, heuristic scoring criteria. The proposed algorithm's performance is demonstrated by exploring a challenging 3D maze environment and compared with a state-of-the-art algorithm.

## 1 Introduction

Unmanned Aerial Vehicles (UAVs) have received much attention in recent years due to its wide range of applications, such as autonomous structural inspection [1, 2] and exploration of an unknown environment [3-18]. In particular, exploration with the UAV is one of the most popular applications to acquire 3D map of an unknown environment without prior knowledge of it. For the exploration, mapping quality, precise localization, motion planning, and control are mandatory [3]. However, existing methods have been largely challenged by computationally heavy probabilistic path planning [20, 21], which even yields unnecessary back-and-forth movements problem [19]. Moreover, many preceding methods had limitations that prior map information or human-guidance [11, 12] are required. Similarly, kinodynamic constraints or proper sensors considering the payload for UAVs were not considered.

In this paper, to solve those issues and to consider the limited payload and computational resource of UAVs, we propose "Peacock Exploration": A lightweight exploration method for UAV using precomputed minimum snap trajectories [22] which look like a peacock's tail as shown in Fig. 1. Using the widely known, control efficient minimum snap trajectories and OctoMap [23], the UAV equipped with a RGB-D camera can explore unknown 3D environments without any prior knowledge or human-guidance with only $O(\log N)$ computational complexity. It also adopts the Receding Horizon [4, 5, 9, 17] approach and simple, heuristic scoring criteria to guarantee strict avoidance from obstacle and quality of 3D map. The performance of the proposed algorithm is demonstrated by exploring a challenging 3D maze environment which realized under a Gazebo simulator using the Robot Operating System (ROS) [24].

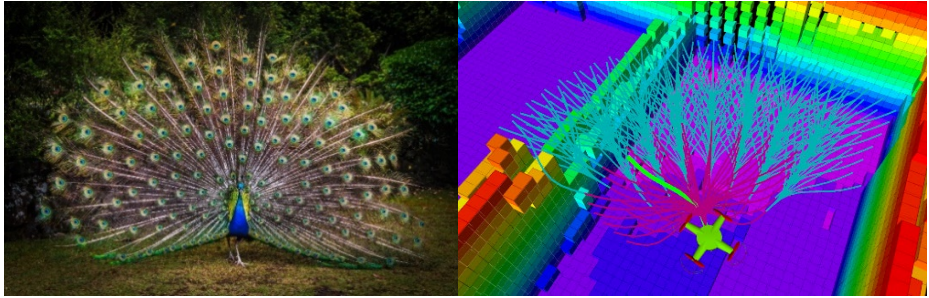

**Fig. 1.** A peacock [25] and peacock trajectory of the proposed algorithm. The colored map represents 3D occupied map from OctoMap [23]. The candidates of receding horizon trajectories are shown in magenta, and the second step trajectories are shown in turquoise, respectively. The best-scored trajectory is shown in the green-thick line.

## 2   Related Work

The purpose of exploration is to map an environment without prior knowledge. As defined in [3], mapping quality, precise localization, motion planning for fast acquisition of information, and safe controller are essential for the exploration. Therefore, recent research on exploration mainly focused on them. Additionally, the research to lighten the computational complexity of the exploration algorithm considering the low computational ability of the UAV is also one of the trends [18].

Research to improve the mapping quality has been conducted by using the concept of the receding horizon [4, 5] or by using coverage path planning [6, 7]. In the study of designing trajectory in consideration of navigation, the uncertainty of localization itself can be reduced by revisiting the features that were previously used in visual localization to see features with low uncertainty [8, 9]. For the same purpose, information-based path planner [10] was used.

To design the safe and fast controller, several trajectory candidates considering the dynamics of the robot are generated [11, 12]. Similarly, research to reduce control complexity by simplifying control points in the guaranteed C-space area was conducted [13]. Moreover, when it is determined that there are no more areas to be mapped in the local area, a study that creates a new path to an appropriate location using the existing path stored in the graph structure has been published [14]. Research that improves both maps and poses uncertainty through machine learning [15, 16] or hierarchical path planning [17] are state-of-the-art methods to consider all four characteristics.

In this paper, to apply exploration to the UAV, the map expressed in the form of OctoMap [23] was used to enable fast path collision inspection. As pointed out by the author of [19], the probabilistic path planning algorithms, such as RRT [20] and PRM [21], show the lousy performance on space visiting efficiency. By creating a minimum snap trajectory bunch called the local path candidate group, the spatial visiting inefficiency has been reduced compared to the probability-based path.

## 3 Methodology

### 3.1 Peacock Trajectory: A Bunch of Minimum Snap Trajectories

The minimum snap trajectory [22] was developed to consider the efficiency of control for a quadrotor and has been widely used in many fields [26, 27] due to its low computational complexity and the ability to consider kinodynamic constraints. The minimum snap trajectory can be obtained by optimizing the cost function as follows:

$$\begin{cases} x^*(t) = \underset{x(t)}{\operatorname{argmin}} \int_0^T \mathcal{L}(x^{(n)}, x^{(n-1)}, \dots, \dot{x}, x, t) dt \\ \qquad\qquad \mathcal{L} = x^{(n)} \end{cases} \quad (1)$$

where $x(t)$ denotes the position function of time, and $\mathcal{L}$ denotes the cost function. Since control inputs of the UAV are strongly related to the fourth derivative of position, the snap, the order of the cost function, $n$, is set as four. This function can be minimized by solving the Euler-Lagrange Equation as below:

$$\frac{\partial \mathcal{L}}{\partial p} - \frac{d}{dt}\left(\frac{\partial \mathcal{L}}{\partial \dot{p}}\right) + \frac{d^2}{dt^2}\left(\frac{\partial \mathcal{L}}{\partial \ddot{p}}\right) + \dots + (-1)^n \frac{d^n}{dt^n}\left(\frac{\partial \mathcal{L}}{\partial p^{(n)}}\right) = 0. \quad (2)$$

By solving the Euler-Lagrange Equation for each of X, Y, and Z position, polynomial functions of time can be obtained, which are desired position over time in each axis, as follows:

$$x^*(t) = c_k t^k + c_{k-1} t^{k-1} + \dots + c_1 t + c_0, \ k = 2n - 1. \quad (3)$$

To obtain a bunch of minimum snap trajectories, which is the peacock trajectory, Eq. (3) is used to solve linear systems with the form of $Ax=b$ with the desired position, velocity, acceleration, and jerk constraints at the time $T$. By setting the linear velocity and the ranges of yaw and pitch for search, the constraints are derived as below:

$$x(T) = x(t_0) + vT\cos(\psi_i)\cos(\theta_j), \quad i = 0, 1, \ldots, row, j = 0, 1, \ldots, col$$

$$y(T) = y(t_0) + vT\sin(\psi_i)\cos(\theta_j), \quad i = 0, 1, \ldots, row, j = 0, 1, \ldots, col$$

$$z(T) = z(t_0) + v\sin(\theta_j), \quad j = 0, 1, \ldots, col \qquad (4)$$

where $v$, $T$ are defined as linear velocity and time period each. $\psi_i$ and $\theta_j$ denote the sets of yaw angles and pitch angles in the same $row*col$ dimension with the score matrix. In the following experiment, row and col are set to nine, to cover [-60°, 60°] yaw range and [-40°, 40°] pitch range. In a similar way, the constraints for the second step trajectories from each (*row*, *col*) paired first step trajectories are derived as:

$$x(2T) = x(T) + v\cos(\overline{\psi_b}), \quad b = 0, 1, \ldots, branch$$

$$y(2T) = y(T) + v\sin(\overline{\psi_b}), \quad b = 0, 1, \ldots, branch \qquad (5)$$

where *branch* is set to seven, to cover [-27°, 27°] yaw range from the end point of the first step trajectories. To maximize efficiency of coverage area and remove redundant region, Z axis value is fixed at the second step trajectories. With the constraints derived from above, peacock trajectory can be precomputed offline using C++ or MATLAB before operating algorithm, and the results are described in Fig. 2.

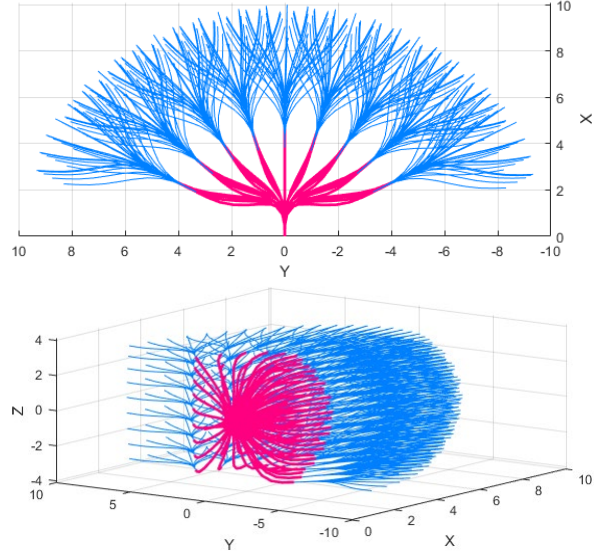

**Fig. 2.** Precomputed peacock trajectory shown in XYZ coordinates in the meter unit. Magenta colored trajectories indicate the first step trajectories, which are candidates of receding horizon trajectory, and blue colored trajectories represent the second step searching trajectories.

Precomputation of peacock trajectory only takes around one millisecond with the experimental computer, which is described in the Appendix. The first step trajectories and the second step trajectories are used to search occupied and unoccupied space for the remaining algorithm. In particular, one of the first step trajectories is selected as a receding horizon trajectory, according to the scoring method in the following section.

### 3.2 Algorithm

Using the precomputed peacock trajectory and OctoMap, the best-scored path is selected according to a simple heuristic policy that gives different weights on free space and unknown space to explore the unknown environment actively. Only $O(\log N)$ computational complexity is required to search $N$ points within the set of precomputed peacock trajectory using octree to lightweight the proposed algorithm. The scoring policy returns *best_row* and *best_col* location of the score matrix, which are the location the best score among the first step trajectories. The selected trajectory is a receding horizon trajectory because the algorithm proposes to track only the first step trajectory. The detailed scoring method is simplified into the pseudo code below.

|    | Algorithm 1 Peacock Exploration – Best Path Selection |
|----|---|
| 1  | **For** *N*                 ← Points in precomputed peacock trajectories |
| 2  |   **If** OctoMap->search(*N*) **then**          >> Known space |
| 3  |     **If** occupied **then**          >> Occupied space |
| 4  |       score(*row, col*) = 0 |
| 5  |     **else**                          >> Free space |
| 6  |       score(*row, col*) += a |
| 7  |   **else**                                      >> Unknown space |
| 8  |     score(*row, col*) += b            >> b>a, user-defined |
| 9  | **End** |
| 10 | Max_score = score->find(&*best_row*, &*best_col*) |
| 11 | **If** size(Max_score)>2 **then**                         >> same score |
| 12 |   *best_row* = medium(*best_rows*) |
| 13 |   *best_col* = medium(*best_cols*) |
| 14 | **Return** *best_row, best_col*          ← location of the score matrix |

Thanks to its simple structure, the proposed algorithm's computational complexity is only $O(\log N)$, and it does not need additional procedures other than precomputation and octree search. The computational complexities compared with the other widely known algorithms are informed in Table 1.

**Table 1.** Computational complexities of path planning algorithms in Big O notation.

| Algorithm | Precomputation | Octree Search | Processing Nodes | Query Nodes |
|---|---|---|---|---|
| RRT  | No | $O(\log N)$ | $O(N\log N)$ | $O(N)$ |
| RRT* | No | $O(\log N)$ | $O(N\log N)$ | $O(N)$ |
| Ours | Yes – avg. 1ms | $O(\log N)$ | Precomputed | No |

### 3.3 Trajectory Tracking Controller

To track the selected best-scored receding horizon trajectory precisely, an accurate nonlinear controller, Geometric Tracking Controller [28], is adopted here. It guarantees almost globally exponential stability on the Special Euclidean group (SE(3)) without any chance of singularities. Body-fixed angular velocities and throttle inputs can be calculated by obtaining the desired z-axis throttle vector as follows:

$$\overrightarrow{b_{z,d}} = \frac{m\ddot{p}_d + k_p e_p + k_v e_v + mge_3}{\|m\ddot{p}_d + k_p e_p + k_v e_v + mge_3\|}$$

Where $e_p$ denotes the errors between the desired position and current position, and $e_v$ indicates the errors between the desired velocity and current velocity, respectively. Here, $k_p$, $k_v$ are gains, and $m$, $g$ stands for the mass of the quadrotor and gravitational acceleration each. The more detailed derivation of inputs is described in [28].

The ground-truth position and velocities are accessible thanks to the simulation environment. The desired position, velocities can be easily acquired from the precomputed trajectory polynomial function of time.

## 4 Experiment

A ROS based Gazebo simulator [24] environment is chosen due to its precise and real-like physics. A customized, challenging 3D maze was built with eight meters tall walls and four meters tall upper blocked or lower blocked walls, as shown in Fig. 3. The detailed dimensions and the parameters are summarized in Table. 2.

The UAV in the simulator is equipped with a RGB-D camera rather than a heavy LiDAR, to consider its payload. The depth information is converted into 3D point cloud data and inserted into the OctoMap. After autonomous take-off of two meters high altitude, the proposed algorithm starts to operate and then tries to complete the map along its selected trajectory. If it arrives at the final position, the algorithm stops, and then the UAV lands. While exploring the environment, volumetric information of unoccupied, occupied, and total searched spaces are recorded to be analyzed. An open-source and state-of-the-art algorithm "MBP" [29], which adopts a global planner and a LiDAR as default, is also executed and compared.

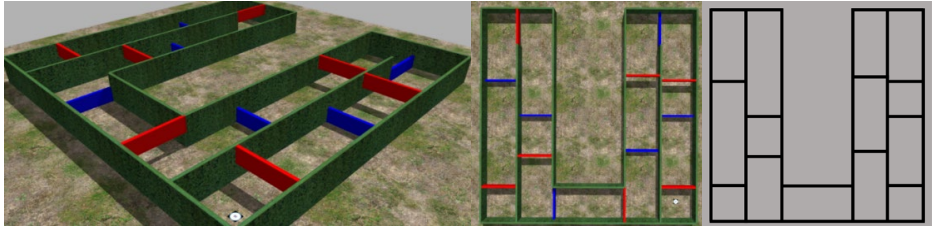

**Fig. 3.** Side-view and Top-view of a customized 3D maze in Gazebo simulator. The far-right figure shows the 2D projection map, which cannot be completed with 2D exploration algorithms.

**Table 2.** Experimental parameters and specifications of sensor.

| Parameter | Value | Parameter | Value |
|---|---|---|---|
| Dimension | 90*90*8m | OctoMap Resolution | 0.5*0.5*0.5m |
| Rows ($\theta_j$) | 9, [-40°, 40°] | OctoMap Hit Probability | 0.65 |
| Columns ($\psi_i$) | 9, [-60°, 60°] | OctoMap Miss Probability | 0.35 |
| Branches ($\overline{\psi_b}$) | 7, [-27°, 27°] | OctoMap Search Depth | 15 level |
| Max Velocity | $v$ = 5m/s (Ours) | Camera Resolution | 640x480 |
|  | 2.5m/s (MBP) | Camera Depth Range | 0.11 – 15m |
| LiDAR Range | 20m (MBP) | Camera FoV | 60° x 45° |

**Table 3.** Results of the experiment.

| Algorithm | Duration | Flight Length | Avg. Velocity | Mapped Volume | Avg. Rate of Mapping | Mapping Efficiency |
|---|---|---|---|---|---|---|
| MBP | 326.90s | 773.642m | 2.37m/s | 11,641.75m$^3$ | 35.61m$^3$/s | 15.05m$^3$/m |
| Ours | 140.24s | 467.327m | 3.33m/s | 27,712.125m$^3$ | 154.82m$^3$/s | 46.46m$^3$/m |

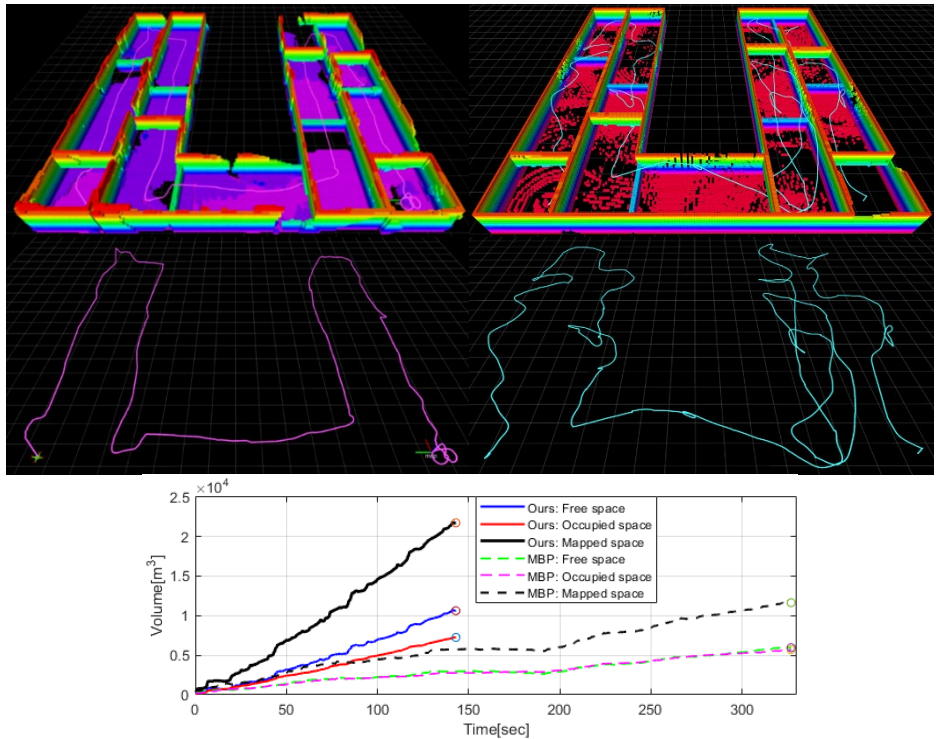

**Fig. 4.** The occupied OctoMaps after the experiment and the path of the UAV. The left and right columns' figures show the results of ours and MBP, respectively. Grid cells represent five meters. The last graph shows the mapped volume versus time, which is the sharper, the better.

## 5      Result and Discussion

The results of the autonomous exploration with the proposed algorithm are shown in Fig. 4. and in Table 3. for both qualitative and quantitative evaluations. The colored maps of the first row show the explored, occupied space according to the maps' height. The long line maps indicate the flight trajectory of the UAV through exploration. The maps for the left and right columns represent the results of ours and MBP, respectively. Additionally, the last graph represents the mapped volume versus time, which is the sharper, the better.

As outlined in the introduction, the proposed peacock exploration method showed good enough performance of mapping according to the results. It completed the challenging 3D maze at the averagely 3.33 meters per second, faster than MBP, only using a single RGB-D camera rather than a LiDAR with a longer sensing range. MBP showed the unnecessary back-and-forth movements and even crashed onto walls when the maximum velocity is set higher than 2.5 meters per second. The average mapping rate showed an overwhelmingly better performance than MBP, even without an additional global planner. In addition, the proposed algorithm tried to not only complete the map but also explore as much as possible, as it can be seen that there were rotating and detouring actions from the flight path. Even with the shorter flight trajectory, the proposed algorithm mapped more spacious volume, which means better mapping efficiency.

## 6      Conclusion

In this paper, we proposed the lightweight exploration method, peacock exploration, for UAVs using control efficient minimum snap trajectories. From the experiment under the challenging 3D environment, it showed good enough mapping quality without an additional global planner but only with lightweight peacock trajectory and the heuristic scoring method. Moreover, thanks to its receding horizon based method, it guarantees the full obstacle avoidance without prior knowledge of the map.

Since it is adopting a smooth minimum snap trajectory as its local planner, the proposed algorithm seems to be able to extend to consider the localization uncertainty in future work. The experiment and the results can be watched as video clip here.
`https://youtu.be/6K-QSb1Aq5E`
`https://youtu.be/t3ysB8Y_GCU`

## 7      Acknowledgement

The students are supported by Korea Ministry of Land, Infrastructure and Transport (MOLIT) as "Innovative Talent Education Program for Smart City" and BK21 FOUR.

# Appendix

The specification of the experimental computer is shown at the table below.

**Table 4.** Experimental Computer Setup

| | |
|---|---|
| CPU | Intel® Core™ i5-9600K @3.70Hz |
| RAM | Samsung DDR4 16GB PC4-19200 * 2 @2400MHz |
| GPU | GeForce® GTX1650 4GB GDDR6 @1530MHz |
| OS | Ubuntu 18.04LTS – ROS Melodic |